\documentclass[11pt,a4paper]{article}
\usepackage[hyperref]{naaclhlt2019}

\usepackage{times}
\usepackage{latexsym}
\usepackage{bbm}
\usepackage{multirow}
\usepackage{multicol}
\usepackage{subfiles}
\usepackage{url}
\usepackage{epsfig}
\usepackage{graphicx}
\usepackage{amsmath}
\usepackage{amssymb}
\usepackage{multirow}
\usepackage{tabularx}
\usepackage{subcaption}
\usepackage{hyperref}

\definecolor{newgreen}{RGB}{156, 193, 135}
\definecolor{newblue}{RGB}{139, 166, 203}
\definecolor{neworange}{RGB}{218, 169, 63}

\DeclareMathOperator*{\argmax}{argmax}
\aclfinalcopy %
\def\aclpaperid{2216} %

\newcommand{\setsymbol}[2]{%
  \expandafter\gdef\csname @icmlsymbol#1\endcsname{#2}
 }

\title{ExCL: Extractive Clip Localization Using Natural Language Descriptions}
\author{Soham Ghosh$^{1,*}$
  \\\And
  Anuva Agarwal$^{1,*}$
  \\\And
  Zarana Parekh$^{1,}$\thanks{~\,Equal contribution, randomly ordered.}
  \\\And
  Alexander Hauptmann$^{1}$ \\
  \AND\\[-5ex]
$^{1}$Language Technologies Institute\\Carnegie Mellon University
\AND\\[-5ex]
    \texttt{\{sohamg,anuvaa,zpp,alex\}@cs.cmu.edu}}

\date{}

\begin{document}
\maketitle

\begin{abstract}
    
The task of retrieving clips within videos based on a given natural language query requires cross-modal reasoning over multiple frames. Prior approaches such as sliding window classifiers are inefficient, while text-clip similarity driven ranking-based approaches such as segment proposal networks are far more complicated. In order to select the most relevant video clip corresponding to the given text description, we propose a novel extractive approach that predicts the start and end frames by leveraging cross-modal interactions between the text and video - this removes the need to retrieve and re-rank multiple proposal segments. Using recurrent networks we encode the two modalities into a joint representation which is then used in different variants of start-end frame predictor networks. Through extensive experimentation and ablative analysis, we demonstrate that our simple and elegant approach significantly outperforms state of the art on two datasets and has comparable performance on a third.

\end{abstract}

\section{Introduction}
    \label{sec:intro}
\begin{figure}
    \centering
    \includegraphics[width=\columnwidth]{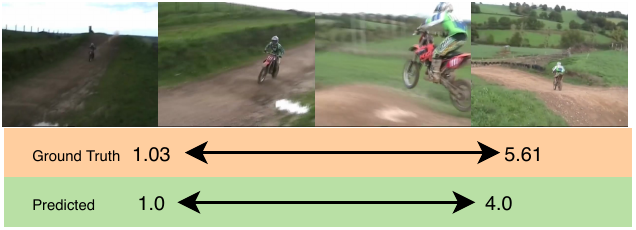}
    \caption{Clip Extraction task for the given query `\textit{the biker jumps to another ramp near the camera}'}
    \label{fig:prediction}
\end{figure}

    Clip Localization is the task of selecting the relevant span of temporal frames in a video corresponding to a natural language description and has recently piqued interest in research that lies at the intersection of visual and textual modalities. 
An example of this task is demonstrated in Figure \ref{fig:prediction}.
It requires cross-modal reasoning to ground free-form text inside the video and calls for models capable of segmenting a video into action segments \cite{singh2016multi, yeung2016end,xu2017r} as well as measuring multi-modal semantic similarity \cite{karpathy2015deep}. 

This task is inherently discriminative, i.e., there is only a single most relevant clip pertaining to a given query in the corresponding video. However, most prior works \cite{hendricks2017localizing, hendricks2018localizing, liu2018temporal, chen2018temporally, zhang2018man} explore this as a ranking task over a fixed number of moments by uniformly sampling clips within a video. Moreover, these approaches are restrictive in scope since they use predefined clips as candidates for a video and cannot be easily extended to videos with considerable variance in length.

\citet{gao2017tall,xu2019multilevel} apply two-stage methods which rank candidate clips using a learned similarity metric. \citet{gao2017tall,ge2019mac} propose a sliding window approach with alignment and offset regression learning objective, but it is limited by the coarseness of the windows and is thus inefficient and inflexible. \citet{xu2019multilevel} address this through a query-guided segment proposal network (QSPN). However, the similarity metric used by these approaches is difficult to learn as it is sensitive to the choice of negative samples \cite{triplet} and it still does not consider the discriminative nature of the task.

Hence, we propose an elegant and fairly simple extractive approach. Our technique is similar to text-based Machine Comprehension \cite{chen2017reading} but in a multimodal setting where the video is analogous to the text passage and the target-clip is analogous to the text span corresponding to the correct answer. We verify empirically that our method significantly outperforms prior work on two benchmark datasets - TACoS, ActivityNet and comparably well on the third, Charades-STA. 
Our flexible, modular approach to \textit{Extractive Clip Localization (ExCL)} can easily be extended to incorporate attention models and different variants of encoders for both visual and text modality to improve performance further.

\section{Approach}
    \label{sec:approach}
    \begin{figure*}[t!]
\centering
    \begin{subfigure}[t]{0.27\textwidth}
        \centering
        \includegraphics[width=\textwidth]{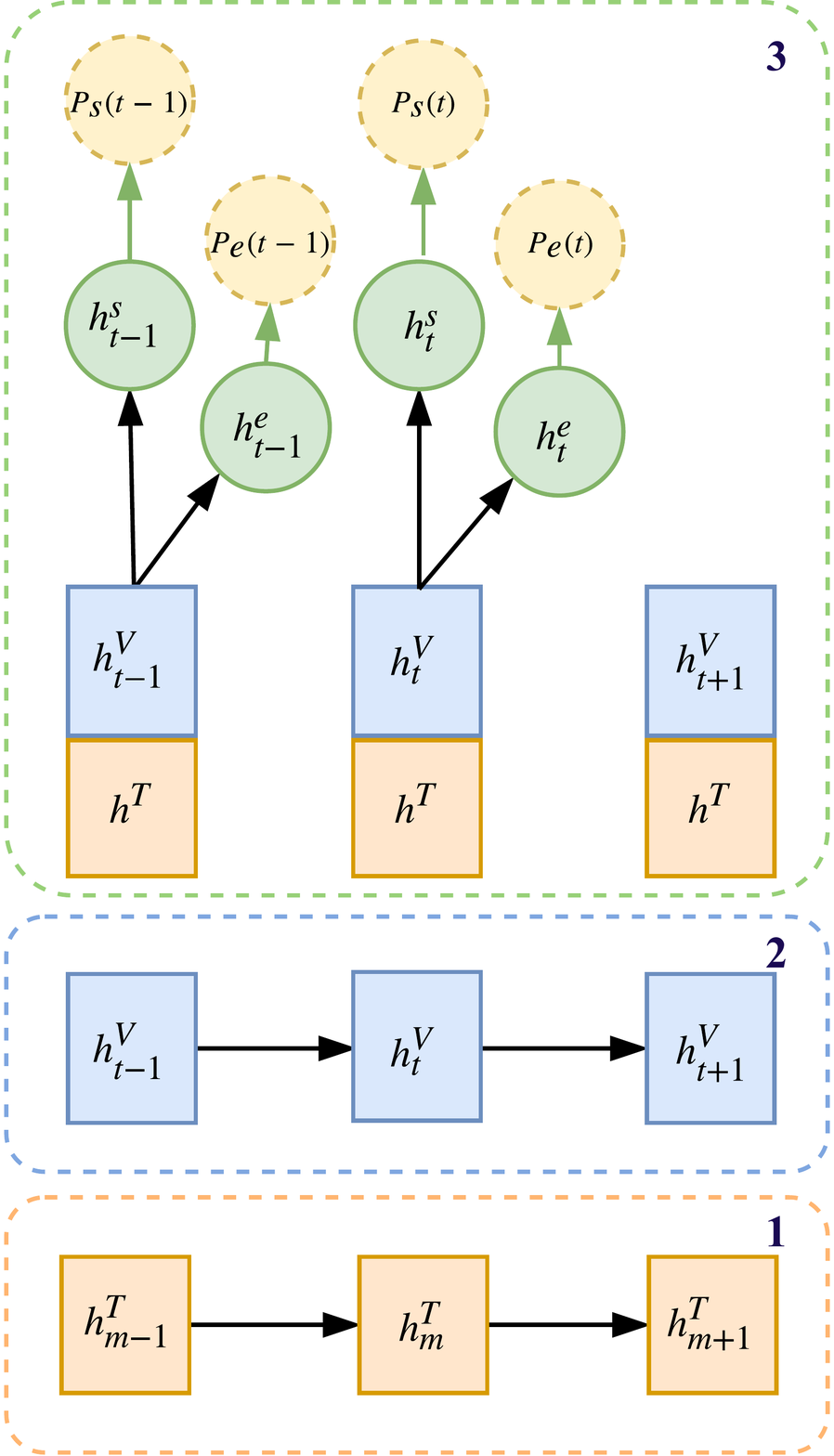}
        \caption{MLP predictor}
    \end{subfigure}
    \begin{subfigure}[t]{0.27\textwidth}
        \centering
        \includegraphics[width=\textwidth]{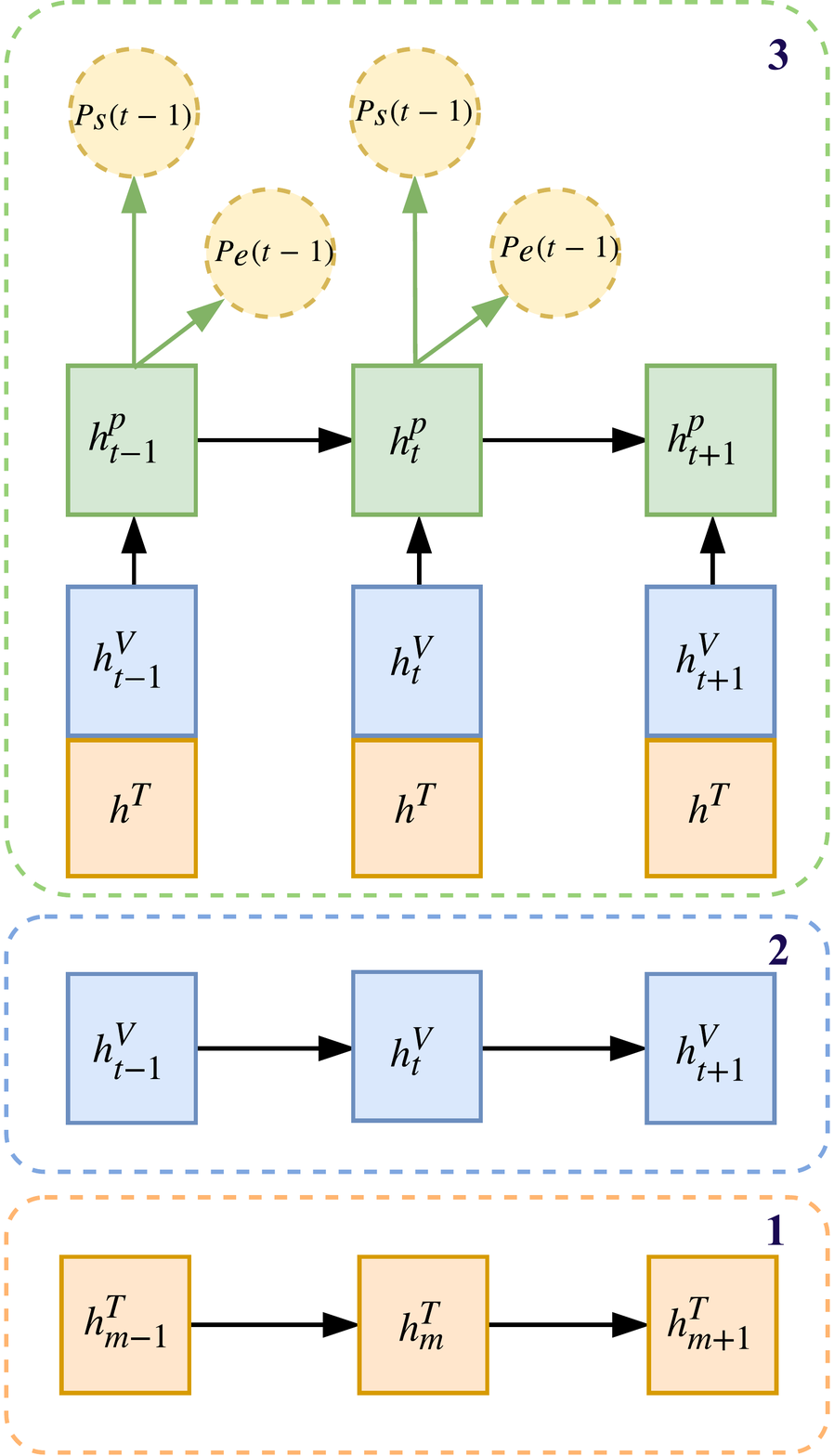}
        \caption{Tied-LSTM predictor}
    \end{subfigure}
    \begin{subfigure}[t]{0.27\textwidth}
        \centering
        \includegraphics[width=\textwidth]{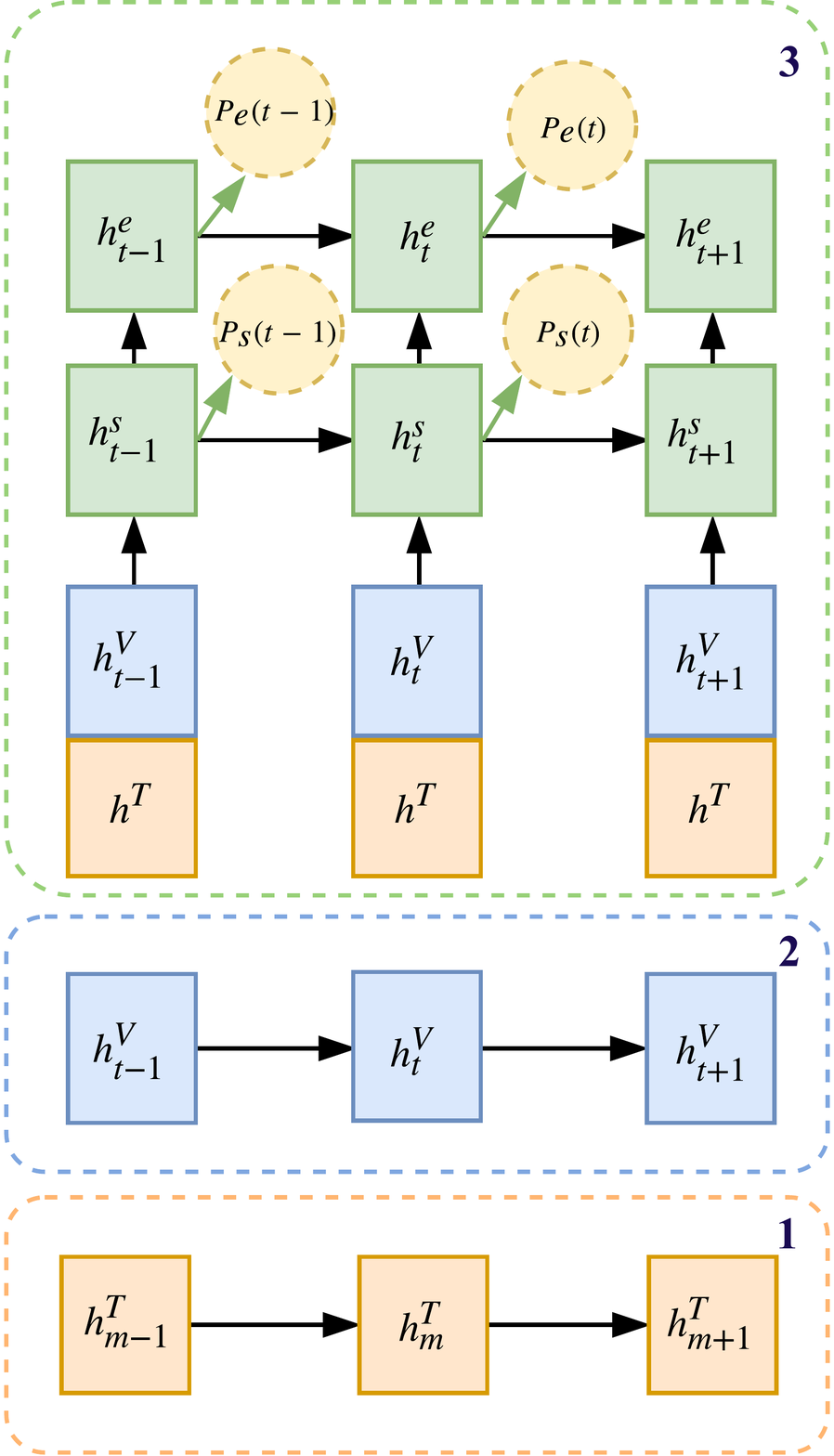}
        \caption{Conditioned-LSTM predictor}
    \end{subfigure}
    \caption{Our model consists of three modules: a text sentence encoder (\textbf{\textcolor{neworange}{orange}}, denoted by $[1]$), a video encoder (\textbf{\textcolor{newblue}{blue}}, denoted by $[2]$) and three variants of span predictor (\textbf{\textcolor{newgreen}{green}}, denoted by $[3]$) - MLP, Tied-LSTM and Conditioned-LSTM to predict start and end probabilities for each frame. We use bidirectional LSTMs for the text and video encoders. }\label{fig:model}
\end{figure*}

Our model comprises of three modular parts - a text encoder, a video encoder and a span predictor  as shown in Figure \ref{fig:model}. We use a bidirectional text LSTM as the text encoder with pre-trained GloVE \cite{glove} embeddings as input, and use the last hidden state ($\mathbf{h}^T$) as sentence-embedding. Simpler variants such as bag-of-words, and other approaches such as InferSent \cite{infersent} or Skip-Thought \cite{skipthought} could also be used. We use a bidirectional LSTM with I3D features \cite{carreira2017quo} as the video encoder which captures temporal context at each time-step ($\mathbf{h}^V_t$). Note that this could also be substituted by C3D features \cite{c3d} or self-attentive networks \cite{selfattn_vqa}. Finally, we compare the variants of span-predictor networks which output scores $S_{\text{start}}(t), S_{\text{end}}(t) \in \mathbb{R}^1$ respectively. 

\subsection{Training objectives}
We consider two modes of training our networks, one which uses a classification loss (ExCL-clf) and another which uses a regression loss (ExCL-reg).
\paragraph{\textbf{ExCL-clf:}} The scores  are normalized using SoftMax to give $P_{\text{start}}(t), P_{\text{end}}(t)$, and trained using negative log-likelihood loss:
\begin{align}
    L(\theta) &= - \frac{1}{N} \sum_i^N \log(P_{\text{start}}(t^i_s)) + \log(P_{\text{end}}(t^i_e))
\end{align}
where $N$ is number of text-clip pairs and $t^i_s, t^i_e$ are the ground-truth start and end frame indices for the $i^{th}$ pair. During inference we predict the span of frames ($\hat{t}_s, \hat{t}_e$) for each query by maximizing the joint probability of start and end frames.
\begin{align}
 \operatorname{span}(\hat{t}_s, \hat{t}_e) &= \argmax_{\hat{t}_s,\hat{t}_e} P_{\text{start}}(\hat{t}_s)P_{\text{end}}(\hat{t}_e)  \\
  &= \argmax_{\hat{t}_s,\hat{t}_e} S_{\text{start}}(\hat{t}_s) + S_{\text{end}}(\hat{t}_e) \\
  & \text{ s.t. } \hat{t}_s \leq \hat{t}_e
\end{align}

\paragraph{\textbf{ExCL-reg:}} The classification approach is limited to predicting start and end times of clip at discretized intervals, while the true target is a continuous value. In order to directly model this as a regression problem, we formulate start and end time prediction by computing an expectation over the probability distribution given by SoftMax outputs:
\begin{align} 
    t_s &= \mathbb{E}_{P_\text{start}}\big[ t \big] \\
    &= \sum_{t_s=1}^T  t_s P_\text{start}(t_s)
\end{align}
\begin{align}\label{eq:t-end-excl-reg}
    t_e &= \mathbb{E}_{P_\text{start}}[\mathbb{E}_{P_{\text{end}|\text{start}}}[t]] \\ 
    &= \sum_{t_s=1}^T P_\text{start}(t_s)\sum_{t_e=1}^T P_{\text{end}|\text{start}}(t_e)
\end{align}
Here the above equations $t_s, t_e$ refer to the actual time values corresponding to each index. $P_{\text{end}|\text{start}}(t_e)$ is computed by a SoftMax over masked logits:
\begin{align*}
    P_{\text{end}|\text{start}} &= \operatorname{SoftMax}(\mathbbm{1}[t_e \geq
    t_s]S_{\text{end}}(t))
\end{align*}
Finally we train the networks using regression losses such as mean squared error and absolute error. We find that using absolute error and first normalizing the values of time to $t_s, t_e \in [0,1]$ yields better results.

\subsection{Span Predictor Variants}
We implement the following variants of the span predictor network:
\paragraph{MLP predictor:} \label{ssec:mlp} At each time step $t$, we pass concatenated video-encoder features with sentence embeddings into two multi-layered perceptrons (MLPs) to obtain scores $S_{\text{start}}(t), S_{\text{end}}(t)$. 
\begin{align}
    S_{\text{start}}(t) &= \operatorname{MLP}_{\text{start}}([\mathbf{h}^V_t; \mathbf{h}^T])\\
    S_{\text{end}}(t) &= \operatorname{MLP}_{\text{end}}([\mathbf{h}^V_t; \mathbf{h}^T]) 
\end{align}
\paragraph{Tied LSTM predictor:}\label{ssec:tied-lstm} Here, we send concatenated video encoder output and sentence embedding to a bidirectional LSTM as 
input at each step in order to capture recurrent cross-modal interactions. The hidden states ($\mathbf{h}^P_t$) concatenated with the original inputs are then fed to a MLP with $\tanh$ activation in hidden layers to predict start and end scores.
\begin{align}
    h^P_t &= \operatorname{LSTM}([\mathbf{h}^V_t; \mathbf{h}^T], \mathbf{h}^P_{t-1}) \\
    S_{\text{start}}(t) &= \operatorname{MLP}_{\text{start}}([\mathbf{h}^P_t; \mathbf{h}^V_t; \mathbf{h}^T])\\
    S_{\text{end}}(t) &= \operatorname{MLP}_{\text{end}}([\mathbf{h}^P_t; \mathbf{h}^V_t; \mathbf{h}^T])
\end{align}
\paragraph{Conditioned LSTM predictor:}\label{sec:cond-lstm} Note that in the previous two approaches the end-frame predictor is not conditioned in any way on the start predictor.
Here we use two bidirectional LSTMs: $\operatorname{LSTM}_{\text{end}}$ takes as input the hidden states $\mathbf{h}^{P_0}_t$ of $\operatorname{LSTM}_{\text{start}}$ and produces $\mathbf{h}^{P_1}_t$ as output.
The respective hidden states are then used in a similar way as the tied LSTM method to generate start and end scores.
\begin{align}
    \mathbf{h}^{P_0}_t &= \operatorname{LSTM}_{\text{start}}([\mathbf{h}^V_t; \mathbf{h}^T], \mathbf{h}^{P_0}_{t-1}) \\
    \mathbf{h}^{P_1}_t &= \operatorname{LSTM}_{\text{end}}(\mathbf{h}^{P_0}_t, \mathbf{h}^{P_1}_{t-1}) \\
    S_{\text{start}}(t) &= \mathbf{W}_s ([\mathbf{h}^{P_0}_t; \mathbf{h}^V_t; \mathbf{h}^T]) + \mathbf{b}_s\\
    S_{\text{end}}(t) &= \mathbf{W}_e ([\mathbf{h}^{P_1}_t; \mathbf{h}^V_t; \mathbf{h}^T]) + \mathbf{b}_e
\end{align}

\section{Datasets}
    \label{sec:datasets}

We evaluate our models on three datasets. Note that we do not evaluate our models on DiDeMo/TEMPO  \cite{hendricks2017localizing,hendricks2018localizing} as these datasets only provide coarse fixed-size moments, thus reducing the problem essentially to ranking a fixed set of candidates. We choose the following datasets because each has unique properties such as visual variance, richness of vocabulary and variance in query lengths.

\paragraph{MPII TACoS:} This dataset \cite{rohrbach2014coherent} has been built on top of the MPII Cooking Activities dataset. It consists of detailed temporally aligned text descriptions of cooking activities. The average length of videos is 5 minutes. A significant challenge in TACoS dataset is that descriptions span over only a few seconds because of the atomic nature of queries such as `takes out the knife' and `chops the onion' (8.4\% of them are less than 1.6s long). Such short queries allow a smaller margin of error. We use the train/test split as provided by \citet{gao2017tall} \href{https://drive.google.com/file/d/1HF-hNFPvLrHwI5O7YvYKZWTeTxC5Mg1K/view?usp=sharing}{\textit{here}}. Coupled with lesser visual variance to distinguish activities, fine-grained actions and descriptions which can often have high word overlap, this is a challenging dataset.

\paragraph{ActivityNet Captions:} ActivityNet \cite{caba2015activitynet} is a large-scale open domain activity recognition, segmentation and prediction dataset based on YouTube videos additionally augmented with dense temporally annotated captions \cite{krishna2017dense}. The average length is 2 minutes, but they are much more diverse in content, with videos that span over 200 activity classes and annotations which use a richer vocabulary. There are 10,024 and 5,044 train and test set videos. We use the same train/test split as provided by the authors. 
Our reported results have 3,370 missing videos which could not be downloaded.

\paragraph{Charades-STA:} The Charades dataset was introduced with action classification, localization and video description annotations \cite{sigurdsson2016hollywood}. \citet{gao2017tall} extend this to include sentence level temporal annotation to create the Charades-STA dataset which contains 12,408 training, and 3,720 test sentence level annotations. In Charades-STA, average length of videos is ~30 seconds and query length (number of frames) has much lower variance (3.7s) as compared to TACoS (39.5s) and ActivityNet (78.1s).

\begin{table*}[!ht]
\begin{centering}
\begin{tabular}{l|ccc|ccc|ccc}
 & \multicolumn{3}{c|}{TACoS} & \multicolumn{3}{c|}{Charades-STA} & \multicolumn{3}{c}{ActivityNet} \\
IoU & 0.3 & 0.5 & 0.7 & 0.3 & 0.5 & 0.7 & 0.3 & 0.5 & 0.7 \\ \hline
\citet{ge2019mac} & 24.2 & 20.0 & -- & -- & 30.5 & 12.2 & -- & -- & -- \\
\citet{xu2019multilevel} & -- & -- & -- & 54.7 & 35.6 & 15.8 & 45.3 & 27.7 & 13.6 \\
\citet{zhang2018man} & -- & -- & -- & -- & 46.5 & 22.7 & -- & -- & -- \\
\hline
ExCL-clf 1-a & 22.6 & 12.6 & 5.1 & 55.4 & 30.4 & 14.8 & 42.5 & 23.8 & 12.1 \\
ExCL-clf 1-b & 42.0 & 25.0 & 12.3 & 64.7 & 43.8 & 22.1 & 61.7 & 40.4 & 23.0 \\
ExCL-clf 1-c & 41.9 & 25.5 & 13.6 & 64.2 & 43.9 & \textbf{23.3} & 60.7 & 40.9 & 23.4 \\
ExCL-clf 2-a & 41.7 & 26.0 & 12.9 & 64.6 & 41.5 & 20.3 & 60.4 & 40.5 & 23.1 \\
ExCL-clf 2-b & 44.2 & \textbf{28.0} & \textbf{14.6} & \textbf{65.1} & \textbf{44.1} & 22.6 & 61.1 & 41.3 & 23.4 \\
ExCL-clf 2-c & \textbf{44.4} & 27.8 & \textbf{14.6} & 61.4 & 41.2 & 21.3 & \textbf{62.1} & \textbf{41.6} & \textbf{23.9} \\
\hline
ExCL-reg 1-a & 26.2 & 11.9 & 4.8 & 54.7 & 34.0 & 14.5 & 48.4 & 27.0 & 11.0 \\
ExCL-reg 1-b & 45.2 & 27.5 & 12.9 & 60.1 & 42.6 & 21.6 & \textbf{63.0} & \textbf{43.6} & 23.6 \\
ExCL-reg 1-c & 41.4 & 24.8 & 11.4 & 59.0 & 43.1 & 20.7 & 61.5 & 42.7 & 23.4 \\
ExCL-reg 2-a & 42.2 & 27.2 & 11.7 & 59.6 & 41.9 & 20.2 & 61.5 & 41.9 & 23.3 \\
ExCL-reg 2-b & \textbf{45.5} & \textbf{28.0} & \textbf{13.8} & \textbf{61.5} & \textbf{44.1} & \textbf{22.4} & 62.3 & 42.7 & \textbf{24.1} \\
ExCL-reg 2-c & 42.3 & 27.3 & 12.5 & 58.0 & 41.8 & 20.5 & 61.4 & 41.7 & 22.4
\end{tabular}

\caption{Clip Localization Accuracy at IoU = \{0.3, 0.5, 0.7\} for TACoS, Charades-STA and ActivityNet. Here ExCL-clf represents the classification loss model and ExCL-reg represents the regression loss model. ExCL-\{clf/reg\} 1-$m$ refer to models run without a video LSTM encoder, while ExCL-\{clf/reg\} 2-$m$ include the video LSTM. $m = a, b, c$ refer to MLP, tied LSTM and conditioned LSTM span predictor networks respectively.}\label{tab:results}
\end{centering}
\end{table*}

\section{Experiments}
    \label{sec:exps}

\subsection{Feature Extraction}
We downsample the videos at a frame rate of 5 frames per second and extract I3D RGB \cite{carreira2017quo} visual features pretrained on Kinetics dataset. We use a fine-tuned version for Charades available \href{https://github.com/piergiaj/pytorch-i3d/blob/master/models/rgb_charades.pt}{\textit{here}}. 

\subsection{Implementation Details}

We use pre-trained GloVE embeddings of 300 dimensions. The extracted visual features have 1024 dimensions. For TACoS, Charades-STA, and ActivityNet we use a vocabulary size of 1438, 3720 and 10,000 respectively. We considered batch sizes of 16, 32, 64 and single-layer LSTM hidden sizes of 128, 256, 512. The video and span predictors are bidirectional LSTMs with 256 and 128 hidden units respectively while queries are encoded by a 256-dimensional bidirectional LSTM. The MLP based span predictors have 256 hidden dimensions and use $\tanh$ activation in hidden layers. For all datasets, we train the model with a batch size of 32 for 30 epochs using Adam optimizer with a learning rate of 0.001 and early stopping. We apply a dropout of 0.5 to all the above-mentioned LSTMs during training. We measure our model performance primarily using \textit{localization accuracy} which is defined to be intersection over union (IoU) at threshold values of 0.3, 0.5, 0.7 to compare with past work which reports Recall@1, IoU=\{0.3, 0.5, 0.7\}.

\subsection{Experimentation and Ablative Analysis}
We compare the different training objectives (labeled ExCL-clf for classification and ExCL-reg for regression) and evaluate the usefulness of recurrent encoders for video representations by removing the video LSTM (labeled ExCL-{clf/reg} 1-$\{a,b,c\}$). We also perform ablative analysis of a range of span predictor networks. We compare the performance of our proposed model with three baselines which are the current SOTA for the different datsets. 
(i) Activity Concepts based Localizer (ACL) proposed in \citet{ge2019mac} for TACoS (ii) Segment Proposal Network approach proposed in \cite{xu2019multilevel} for ActivityNet and (iii) Moment Alignment Network (MAN) via Iterative Graph Adjustment approach proposed in \cite{zhang2018man} for Charades-STA. %
While we significantly beat the first two baselines for TACoS and Activity Net respectively, we attain comparable performance with the third for Charades-STA. It is to be noted that since the approach in \cite{zhang2018man} depends on ranking fixed number of segments in each video, it is not scalable to the other two datasets which have longer videos with greater variance in their lengths.

\subsection{Results}
Our results in Table \ref{tab:results} confirm our hypothesis that extractive models work better.

We find that across all datasets, models without recurrent architectures (ExCL 1-$a$) perform significantly worse, demonstrating the importance of temporal context provided by LSTMs in either video encoder (ExCL 2-$a$) or span predictors (ExCL 1-$b$). With ExCL 2-$a$ we see that if the visual context is captured well using recurrent architectures in the video LSTM, then even with a MLP span predictor we get a significant improvement in performance.
Furthermore, if we add LSTM span predictors along with the video LSTM (ExCL 2-$\{b,c\}$) we obtain an additional boost in performance.
However, without a recurrent visual encoder, a recurrent span predictor is essential to capture both uni-modal and cross-modal interactions (ExCL 1-\{$b$, $c$\}). 
When comparing the regression learning objective with classification, we do not notice a substantial gain in performance thereby indicating that not much information is lost if the continuous nature of the labels is not considered.
In terms of span predictors, tied LSTM generally performs well across all datasets, and any difference with conditioned LSTM is negligible. This benefit is more pronounced when using the regression objective, possibly because conditioning is already captured in the formulation as given by Equation \ref{eq:t-end-excl-reg}.

While \citet{xu2019multilevel} note a significant difference in performance between Charades-STA and ActivityNet captions, performance is similar on both the datasets in our model. We hypothesize that their model fails to work well when there is large variability in the query lengths, as explained in Section \ref{sec:datasets}.
We also find TACoS to be a significantly more challenging benchmark, similar to \citet{ge2019mac}. %

\section{Conclusion}
    \label{sec:conclusion}
    
In conclusion, our main contribution is an extractive model for clip localization based on text queries as opposed to ranking driven approaches used in the past.
Our results show that this elegant and fairly simple approach works much better empirically on three very different benchmark datasets, with tied LSTM span predictor generally giving best results. It is to be noted that these datasets previously had three different architectures as their respective SOTA, and our work naturally lays the foundation for training a single generalizable model across datasets and possibly related tasks as the next step.
Furthermore, our approach is modular, making it trivial to insert different architectures for the encoders and span-predictors. Other future directions of this work include adding temporal attention in order to handle more complicated temporal references and extending this approach to work for longer, and thereby more challenging videos such as movies.

\section*{Acknowledgements}
The authors would also like to thank anonymous reviewers for their comments and Junwei Liang for helpful discussions. This research was supported in part by DARPA grant FA8750-18-2-0018 funded under the AIDA program. This work is also supported in part through the financial assistance award 60NANB17D156 from U.S. Department of Commerce, National Institute of Standards and Technology and by the Intelligence Advanced Research Projects Activity (IARPA) via Department of Interior/Interior Business Center (DOI/IBC) contract number D17PC00340. 
    
\newpage
\bibliography{naaclhlt2019}
\bibliographystyle{acl_natbib}

\end{document}